# Indexing Irises by Intrinsic Dimension

J. Michael Rozmus
Senior Consultant
rozmus@ieee.org
IEEE Life Senior Member

## Abstract

*28,000+ high-quality iris images of 1350 distinct eyes from 650+ different individuals from a relatively diverse university town population were collected. A small defined unobstructed portion of the normalized iris image is selected as a key portion for quickly identifying an unknown individual when submitting an iris image to be matched to a database of enrolled irises of the 1350 distinct eyes. The intrinsic dimension of a set of these key portions of the 1350 enrolled irises is measured to be about four (4). This set is mapped to a four-dimensional intrinsic space by principal components analysis (PCA). When an iris image is presented to the iris database for identification, the search begins in the neighborhood of the location of its key portion in the 4D intrinsic space, typically finding a correct identifying match after comparison to only a few percent of the database.*

## 1. Introduction

In 1993 [1], John Daugman described his very successful IrisCode method of iris biometric identification by noting the presence or absence of iris features extracted by spatial filtering of normalized iris images with wavelet kernels. The result is a 2048-bit binary feature vector paired with a 2048-bit mask indicating whether there is a sufficiently clear view of each detected feature to use it in a comparison (see section III in [3]).

Iris recognition for the purpose of identifying an unknown individual is typically accomplished today in two major steps: (1) computing an IrisCode (or similar feature vector) and (2) searching for a good match between this feature vector and the feature vectors for a large number of irises enrolled in a database. For very large iris databases like India's AADHAR database of 1.2 billion+ individuals, the searching is practical only by indexing the database so that comparisons to only a modest proportion of irises enrolled in the database is required.

IrisCode and all of the related methods of encoding iris patterns that dominate commercial iris recognition today have proven to be very robust in the face of contrast and brightness variations, variation in ambient illumination, partial occlusion, and other types of "noise". But mapping of iris images into the IrisCode feature vector space does not preserve neighborhood similarity. In other words, images from different eyes that look similar in appearance do not cluster together in IrisCode space according to any known distance measure. This has made indexing a large database of IrisCodes a mathematically difficult problem.

Arora et al present a recent solution to this problem in [2]. Their introduction summarizes previous iris indexing work, which comprises a wide range of sophisticated methods. A few of them cluster the IrisCodes in various ways. Most of them extract from iris images feature vectors different from IrisCode that preserve a useful degree of neighborhood similarity in the alternative feature vector space. Arora et al [2] built a Siamese neural network that maps normalized iris images to 1024-bit binary feature vectors. These vectors are clustered into hundreds of classes, which serve as an index into the iris database. They present evidence that this method is an improvement over previous solutions.

What follows is a novel and much simpler method for indexing irises. A defined unobstructed portion of a normalized iris image is selected as a key portion to narrow the search space for a particular iris in a large database of enrolled irises. The intrinsic dimension of a set of key portions of enrolled irises is measured to be about four (4). The key portions of enrolled irises are mapped to a four-dimensional intrinsic space by principal components analysis (PCA). When an iris image is presented to the iris database for identification, the search begins in the neighborhood of the location of its key portion in the 4D intrinsic space, typically finding a correct identifying match after comparison to only a few percent of the database.

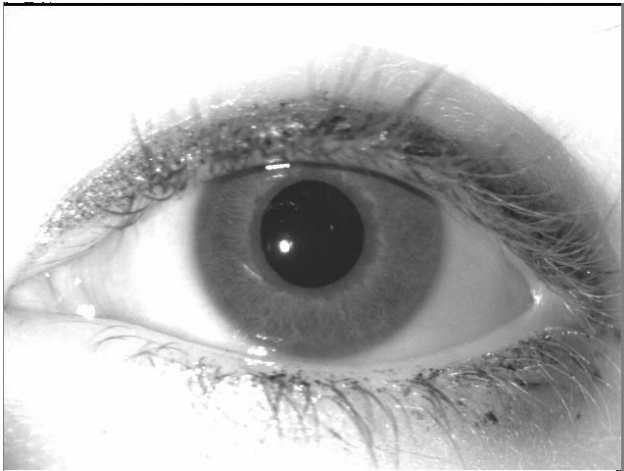
Figure 1: High-quality iris image sample (courtesy of Notre Dame University, CVRL [4])

## 2. High-Quality Iris Images

This research begins with a selection of high-quality iris images from the ND CrossSensor-Iris-2013 Data Set [4]:

- 28,000+ high-quality iris images of 1350 distinct eyes from 650+ different individuals from a relatively diverse university town population captured with the LG4000 iris sensor
- High resolution: iris diameter is about 200-300 pixels (exceeding latest ISO standards for high quality iris images – page 19 in [5])
- Apparently excellent sharpness and signal to noise ratio (see sample in Figure 1)

The original dataset comprised about 30,000 LG4000 images. A Python program was used to discard images with inconsistent metadata and/or file names. Visual inspection corrected two images that were labeled with the wrong subject identification. Visual inspection also removed 11 eye images having opaque contact lenses. The resulting count is 29,792 eye images, which were filtered by more quality measures during segmentation and normalization.

## 3. Development Environment and Tools

All of the research reported here was done using Python 3.9.12 in JupyterLab 3.4.4 running on a robust gaming laptop, an MSI GE76 Raider 11UE, having a Windows 11 OS with 32 GB of RAM. Two sets of software tools were particularly helpful:

1. Python open-source research code for segmentation and normalization of iris images by Adam Czajka et al at Notre Dame CVRL [3]
2. Scikit-Dimension: A Python Package for Intrinsic Dimension Estimation by Jonathan Bac et al [6]

## 4. Algorithm Overview

The iris images are processed as follows:

1. The 29,792 eye images are segmented and normalized into 29,024 64 x 512 normalized irises.
2. A 16 x 256 key portion of each normalized iris is selected. (See Figure 3.) This maps back to a C-shaped lower half of an inner annulus of the iris just below the pupil that is likely to be visible in almost all iris images.
3. The 29,024 key portion images are preprocessed to normalize the range of pixel values, discard images marred by specular reflection, discard images with excessive pixel intensity variance, and to mitigate high-frequency noise. 28,749 key portion images of 1350 different eyes remain.
4. Average the key portion images of each of the 1350 different eyes to reduce noise.
5. Measure the intrinsic dimension of the 1350 average key portion images using the correlation dimension algorithm. The intrinsic dimension of this 4096-dimensional (16 x 256 = 4096) set is measured to be about 4.
6. Use the 1350 average key portion images to define a 4-dimensional (4D) space using PCA. Map the 1350 average key portion images into this newly defined 4D space to generate an enrolled intrinsic iris code (IIC) for each of the 1350 different eyes.
7. For each of the 28,749 key portion image samples of 1350 different eyes:
    a. Map to the newly defined 4D space to generate a query sample IIC.
    b. Search in an increasing neighborhood around this query sample IIC until the enrolled IIC from the same eye is found.
    c. Record the proportion of the 1350 average key portion images that needed to be checked before the search is completed.

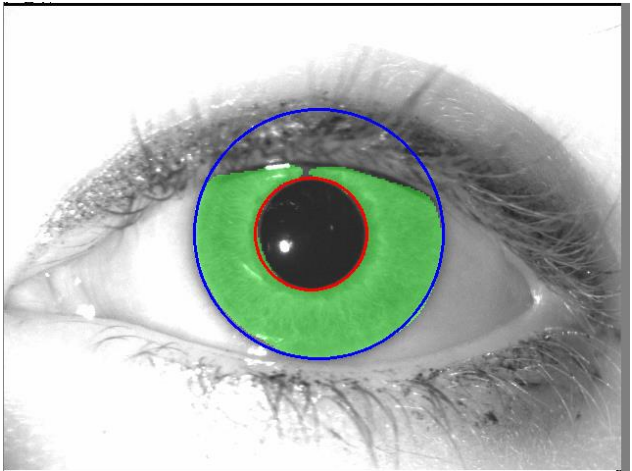

Figure 2: Iris segmentation example of image in Figure 1

### 4.1. Segmentation and Normalization

The Python code from Czajka et al [3] is organized as follows:

1. An efficient fully-trained neural net, called CC-Net [7], finds the iris pixels (green in Figure 2).
2. Circles are fitted to the inner and outer boundaries of the iris pixels (red and blue in Figure 2).
3. A 64 x 512 normalized iris is computed using the rubber sheet model first proposed by John Daugman (page 25 in [8]).

Because the code was originally optimized for forensic investigation using the irises of deceased persons, the following changes were made for the present purpose:

1. Multiple heuristic parameters were tuned to optimize CC-Net for the high-quality images of live people.
2. The radius of the red circle fitting the pupil-iris boundary is increased by 10% of the difference between the blue circle radius and the red circle radius in order to trim away any dark pupil pixels that may get into the normalized iris image due to imperfect fitting.
3. The radius of the blue circle fitting the iris-sclera boundary is decreased by 5% of the difference between the blue circle radius and the red circle radius in order to trim away any bright sclera pixels that may get into the normalized iris image due to imperfect fitting.

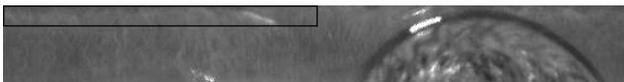

Figure 3: 64 x 512 normalized iris made from the image of Figure 2. The 16 x 256 key portion is outlined in the upper left.

### 4.2. Selection and Preprocessing of Key Portions

Figure 3 shows the normalized iris derived from the same image as Figure 2. The 16 x 256 portion outlined in the upper left is chosen to be used as a kind of key for indexing a large database of irises and their feature vectors. This portion comes from a C-shaped section of the iris just below the pupil in the original eye image that is almost always visible and unobstructed. Although this key portion is imaged with lower resolution than areas of the iris farther from the center of the eye, the results below show that even this relatively low frequency information content is sufficient to serve as an efficient iris index.

The 29,024 key portion images are preprocessed as follows:

1. Normalize the range of pixel values to center on the median value and span +/- 3.5 times the median absolute deviation of the pixel intensity distribution (MAD).
2. Discard normalized irises having more than a few saturated or near-saturated pixels (typically caused by occluding specular reflections).
3. Discard normalized irises having a MAD outside the range of the near infrared reflection pattern of human irises (typically caused by bad segmentation).
4. Mitigate high-frequency noise (shot noise, quantization noise, etc.) by spatial filtering with a 5x5 averaging kernel.

The result is 28,749 key portion images of the 1350 different eyes. An example is shown in Figure 4 below.

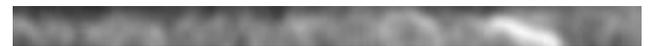

Figure 4: Example of key portion image after preprocessing (derived from normalized image of Figure 3)

In order to minimize the effect of variation in multiple images of the same eye, the average of all key portion images of the same eye is calculated to represent the inherent structure on that portion of each eye. An example of one of the 1350 results from the same eye as Figure 4 is given below in Figure 5.

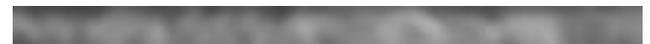

Figure 5: Average of all key portion images of same eye as Figure 4.

### 4.3. Estimation of Intrinsic Dimension

"The intrinsic dimension for a data set can be thought of as the number of variables needed in a minimal representation of the data. Similarly, in signal processing of multidimensional signals, the intrinsic dimension of the signal describes how many variables are needed to generate a good approximation of the signal." [9] In the present work, the complex pattern of the near-infrared reflectivity of the human iris is taken to be a signal communicating a person's identity.

It is well known that noise of any kind increases apparent intrinsic dimension. Consider a set of three-dimensional (3D) points all of which lie in the same plane. The intrinsic dimension of this set is obviously two. But noise will move points out of the common plane. This noise will "thicken" the cloud of 3D points, giving the data set an apparent intrinsic dimension closer to three. Thus, the algorithms described above suppress noise to increase the signal-to-noise ratio of the 1350 distinct human iris signals in the 1350 average key portion images. This enables a good estimation of the intrinsic dimension of these signals.

The correlation dimension algorithm [10] is chosen to estimate the intrinsic dimension of the 4096-dimensional average key portions of the 1350 distinct eyes. In calculating the correlation integral [10], the Scikit-Dimension package specifies the independent distance variable indirectly by number of nearest neighbors [11]. Results are given in Table 1.

Table 1. Correlation Dimension Results

| Smallest Neighborhood Size (% of 1350 samples) | Largest Neighborhood Size (% of 1350 samples) | Estimated Correlation Dimension |
|---|---|---|
| 10 | 20 | 7.92 |
| 20 | 30 | 6.57 |
| 30 | 40 | 5.38 |
| 40 | 50 | 4.43 |
| 50 | 60 | 3.58 |
| 60 | 70 | 2.89 |
| 70 | 80 | 2.24 |
| 80 | 90 | 1.65 |
| 10 | 90 | 4.40 |
| 20 | 80 | 4.18 |
| 30 | 70 | 4.09 |
| 40 | 60 | 4.00 |

The first eight rows of Table 1 give piecewise linear estimates of the slope of a log-log plot of the correlation integral versus neighborhood size. The last four rows fit a line to larger neighborhood ranges centered on 50% of the samples. In the smaller neighborhood ranges, there may not be enough density of samples to correctly represent the geometry of the data set. In the large neighborhood ranges, the calculation of correlation integral may be reaching out past the outer envelope of the whole dataset in 4096-dimensional space. Therefore, a good estimate of the intrinsic dimension is taken to be four (4), corresponding to neighborhood ranges around 50%.

### 4.4. Mapping to Intrinsic Dimension

The 1350 average key portion images are used with the Python package Scikit-Learn to compute a 4-dimensional Intrinsic Iris Code (IIC) space using PCA [12]. Then the 1350 average key portion images are mapped into IIC space to generate an enrolled IIC for each of the 1350 different eyes.

Each of the 28,749 key portion image samples of 1350 different eyes is also mapped into IIC space to generate 28,749 query sample IIC's.

### 4.5. Performance of Indexing with IIC's

Penetration Rate, P, is defined as follows:

$$P = \frac{1}{Q}\sum_{i=1}^{Q} \frac{C_i}{N}$$

Where:

$Q$ is the number of query samples (28,749).
$C_i$ is the number of enrolled IIC's in an expanding neighborhood in IIC space around the IIC of query sample $i$ that must be checked before finding the enrolled IIC that corresponds to the enrolled eye of which the query is a sample.
$N$ is the number of enrolled irises (1350).

Calculation of $P$ proceeds as follows:

1. For each of $Q$ query samples:
    a. Calculate the Euclidean distances in IIC space between the query sample IIC and all $N$ enrolled IIC's.
    b. Sort the pairs of distances and corresponding enrolled iris labels (subject ID plus left/right indicator) according to ascending distance.
    c. Count from the beginning of the sorted pair list until the enrolled iris label matches the query sample iris label. This count is $C_i$.
    d. Calculate a penetration sample $C_i / N$.
2. Average penetration samples of all Q query samples to get $P$.

Figure 6 shows a histogram of the penetration samples. The penetration rate, $P$, calculates to 0.030 (3.0%).

This calculation of penetration rate is a prediction of indexing performance that is done by examining the statistics of all possible $Q$ x $N$ distances in IIC space. In an efficient practical implementation of iris indexing using IIC's, a neighborhood in IIC space of a query sample IIC, having gradually increasing size, would be searched for enrolled IIC's. The feature vector (IrisCode or alternative) of the enrolled iris associated with each enrolled IIC that is found would be checked for a match with the feature vector of the query sample iris. The increasing neighborhood of search in an efficient practical implementation is analogous to step 1c above.

It is also assumed here that the false non-match rate of feature vectors is zero. In other words, when the indexing retrieves the enrolled eye of which the query image is a sample, the match is good and the database search is done.

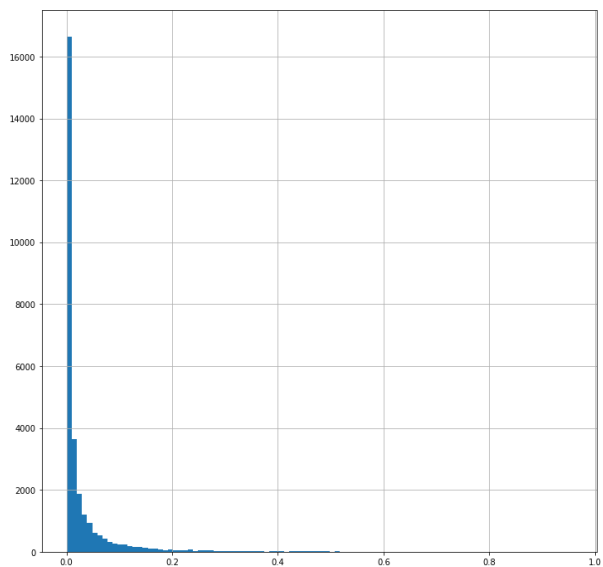

Figure 6: Histogram of penetration of all 28,749 query samples

## 5. Further Research

The intrinsic dimension of a key 16 x 256 portion of normalized iris images is estimated to be four (4). Using linear PCA dimension reduction of these key portion images is an effective means of indexing a large database of irises with four numerical indices. But there is no assurance that linear PCA is an optimal mapping to the intrinsic dimension of four.

Consider Table 2, which gives the calculated penetration rate, $P$, for linear PCA dimension reduction to multiple different dimensions in the neighborhood of 4. The substantial continuing improvement for dimension greater than four suggests that optimal mapping to the measured intrinsic dimension of 4 is non-linear. Non-linear mapping to dimension 4 by a convolutional auto-encoder neural network or a 4D self-organizing map or other non-linear means may improve indexing performance in return for a higher complexity and computation cost.

Table 2. Penetration Rate vs. PCA Mapping Dimension

| PCA Mapping Dimension | Penetration Rate, $P$ |
|---|---|
| 2 | 0.076 |
| 3 | 0.044 |
| 4 | 0.030 |
| 5 | 0.021 |
| 6 | 0.017 |

## References

[1] J. G. Daugman, "High confidence visual recognition of persons by a test of statistical independence," IEEE Trans. Pattern Anal.Mach. Intell., vol. 15, no. 11, pp. 1148–1161, Nov. 1993.

[2] Geetika Arora, Shantanu Vichare, and Kamlesh Tiwari, "IrisIndexNet: Indexing on Iris Databases for Faster Identification", CODS-COMAD 2022, Bangalore, India, https://dl.acm.org/doi/10.1145/3493700.3493715

[3] Adam Czajka, Siamul Karim Khan, and Andrey Kuehlkamp. "Iris recognition designed for post-mortem and diseased eyes", https://github.com/aczajka/iris-recognition---pm-diseased-human-driven-bsif.

[4] ND CrossSensor-Iris-2013 Data Set, Notre Dame University, Computer Vision Research Laboratory (CVRL), https://cvrl.nd.edu/projects/data/#nd-crosssensor-iris-2013-data-set

[5] ISO/IEC 29794-6:2015(E), Information technology — Biometric sample quality — Part 6: Iris Image Data

[6] Bac, J.; Mirkes, E.M.; Gorban A.N., Tyukin, I.; Zinovyev, A., "Scikit-Dimension: A Python Package for Intrinsic Dimension Estimation", Entropy 2021, 23, 1368. https://doi.org/10.3390/e2310136

[7] Suraj Mishra, Adam Czajka, Peixian Liang, Danny Z. Chen, X. Sharon Hu, "CC-Net: Image Complexity Guided Network Compression for Biomedical Image Segmentation," The IEEE Int. Symposium on Biomedical Imaging (ISBI), Venice, Italy, April 8-11, 2019; pre-print available at https://arxiv.org/abs/1901.01578

[8] J. G. Daugman, "How iris recognition works," IEEE Trans. Circuits Syst. Video Technol., vol. 14, no. 1, pp. 21–30, Jan. 2004.

[9] "Intrinsic dimension", Wikipedia, retrieved 08/25/2023

[10] "Correlation dimension", Wikipedia, retrieved 08/25/2023

[11] Scikit-Dimension documentation, skdim.id.CorrInt, https://scikit-dimension.readthedocs.io/en/latest/skdim.id.CorrInt.html, retrieved 08/25/2023

[12] Scikit-Learn documentation, sklearn.decomposition.PCA, https://scikit-learn.org/stable/modules/generated/sklearn.decomposition.PCA.html, retrieved 08/29/2023

**There is a patent pending for the methods described in this paper.**